\documentclass{article}

\usepackage[preprint]{neurips_2025}

\usepackage[utf8]{inputenc} %
\usepackage[T1]{fontenc}    %
\usepackage{hyperref}       %
\usepackage{url}            %
\usepackage{booktabs}       %
\usepackage{amsfonts}       %
\usepackage{nicefrac}       %
\usepackage{microtype}      %
\usepackage{inconsolata}
\usepackage{caption}
\usepackage{amsmath}
\usepackage{amssymb}
\usepackage{mathtools}
\usepackage{enumitem}
\usepackage{makecell} 
\usepackage[usestackEOL]{stackengine}
\usepackage{graphicx}
\usepackage{capt-of}%
\usepackage{booktabs}
\usepackage{varwidth}
\usepackage[table,dvipsnames,svgnames]{xcolor}

\usepackage{tikz}
\usetikzlibrary{shadows}
\usepackage{float}
\usepackage{caption}
\usepackage{subcaption}
\usepackage{xspace}
\usepackage{svg}

\usepackage{import}
\usepackage{animate}
\usepackage{arydshln}
\usepackage{multirow}
\usepackage[normalem]{ulem}
\usepackage[most]{tcolorbox}
\usepackage{colortbl}
\usepackage{tcolorbox}
\usepackage{pifont}
\usepackage{alltt}
\definecolor{titlegray}{rgb}{0.4, 0.4, 0.4} %
\definecolor{contentgray}{rgb}{0.95, 0.95, 0.95} %
\usepackage{algorithm}
\usepackage{algpseudocode}

\newcommand{\model}{\text{Metis-RISE}\xspace}
\newcommand{\modelbf}{\textbf{Metis-RISE}\xspace}

\definecolor{carolinablue}{rgb}{0.6, 0.73, 0.89}
\definecolor{mildgreen}{rgb}{0.85, 0.98, 0.80}
\definecolor{beautycolor}{rgb}{0.91, 0.75, 0.96} %
\definecolor{fallacycolor}{rgb}{0.85, 0.95, 1}
\definecolor{gendercolor}{rgb}{1, 0.85, 0.85}
\definecolor{brightyellow}{RGB}{255, 255, 100}
\definecolor{boxcolor}{RGB}{51,51,153}
\definecolor{lightgreen}{rgb}{0.56, 0.93, 0.56}
\definecolor{citeblue}{HTML}{0064E0}

\hypersetup{
    colorlinks=true,
    citecolor=citeblue,
    linkcolor=red, %
    urlcolor=citeblue %
}

\title{\model: RL Incentivizes and SFT Enhances Multimodal Reasoning Model Learning}

\newtcolorbox{questionbanner}{
  colback=blue!10!white,    %
  colframe=blue!80!black,   %
  width=\textwidth,
  arc=4mm,                  %
  boxrule=1pt,              %
  fonttitle=\bfseries,
  title=Question,
}
\newtcolorbox{promptbox}[1]{
  colback=contentgray,      %
  colframe=titlegray,       %
  colbacktitle=titlegray,   %
  coltitle=white,           %
  title={#1}, %
  arc=4mm,                  %
  rounded corners=northwest, %
  rounded corners=northeast, %
  sharp corners=south,      %
  boxrule=1pt,              %
  fonttitle=\bfseries,      %
}

\definecolor{questionbg}{RGB}{240, 248, 255}  %
\definecolor{answerbg}{RGB}{245, 255, 250}   %
\definecolor{bordercolor}{RGB}{100, 149, 237} %
\definecolor{titlecolor}{RGB}{25, 25, 112}    %

\newtcolorbox{vqaexample}[2][]{
    enhanced,
    breakable,
    colback=white,
    colframe=bordercolor,
    boxrule=1.5pt,
    arc=4pt,
    outer arc=4pt,
    left=8pt,
    right=8pt,
    top=8pt,
    bottom=8pt,
    drop shadow={shadow xshift=0.5mm, shadow yshift=-0.5mm, opacity=0.3},
    overlay={
        \node[
            anchor=north east,
            xshift=-3pt,
            yshift=-3pt,
            fill=bordercolor!80,
            text=white,
            font=\bfseries,
            rounded corners=2pt,
            inner sep=4pt,
            minimum height=1.2em,
            align=center
        ] at (frame.north east) {#2};
    },
    #1
}

\author{Haibo Qiu, Xiaohan Lan, Fanfan Liu, Xiaohu Sun, Delian Ruan, Peng Shi, Lin Ma\thanks{Project Leader.}\\
    \\
    Project Page: \url{https://github.com/MM-Thinking/Metis-RISE} \\
    \\
    \textbf{Meituan} \\
}

\begin{document}

\maketitle

\begin{abstract}
Recent advancements in large language models (LLMs) have witnessed a surge in the development of advanced reasoning paradigms, which are now being integrated into multimodal large language models (MLLMs). However, existing approaches often fall short: methods solely employing reinforcement learning (RL) can struggle with sample inefficiency and activating entirely absent reasoning capabilities, while conventional pipelines that initiate with a cold-start supervised fine-tuning (SFT) phase before RL may restrict the model's exploratory capacity and  face suboptimal convergence. In this work, we introduce \modelbf (\textbf{R}L \textbf{I}ncentivizes and \textbf{S}FT \textbf{E}nhances) for multimodal reasoning model learning. Unlike conventional approaches, \model distinctively omits an initial SFT stage, beginning instead with an RL phase (e.g., using a Group Relative Policy Optimization variant) to incentivize and activate the model's latent reasoning capacity. Subsequently, the targeted SFT stage addresses two key challenges identified during RL: (1) \textit{inefficient trajectory sampling} for tasks where the model possesses but inconsistently applies correct reasoning, which we tackle using self-distilled reasoning trajectories from the RL model itself; and (2) \textit{fundamental capability absence}, which we address by injecting expert-augmented knowledge for prompts where the model entirely fails. This strategic application of RL for incentivization followed by SFT for enhancement forms the core of \model, leading to two versions of our MLLMs (7B and 72B parameters). Evaluations on the \href{https://rank.opencompass.org.cn/leaderboard-multimodal-reasoning/?m=REALTIME}{OpenCompass Multimodal Reasoning Leaderboard}\footnote{Statistics current as of June 26, 2025.} demonstrate that both models achieve state-of-the-art performance among similar-sized models, with the 72B version ranking fourth overall. Please refer to our project page for open-source information.
\end{abstract}

\section{Introduction}
\label{section:introduction}

\begin{figure}[ht]
\centering
\includegraphics[width=1.0\linewidth]{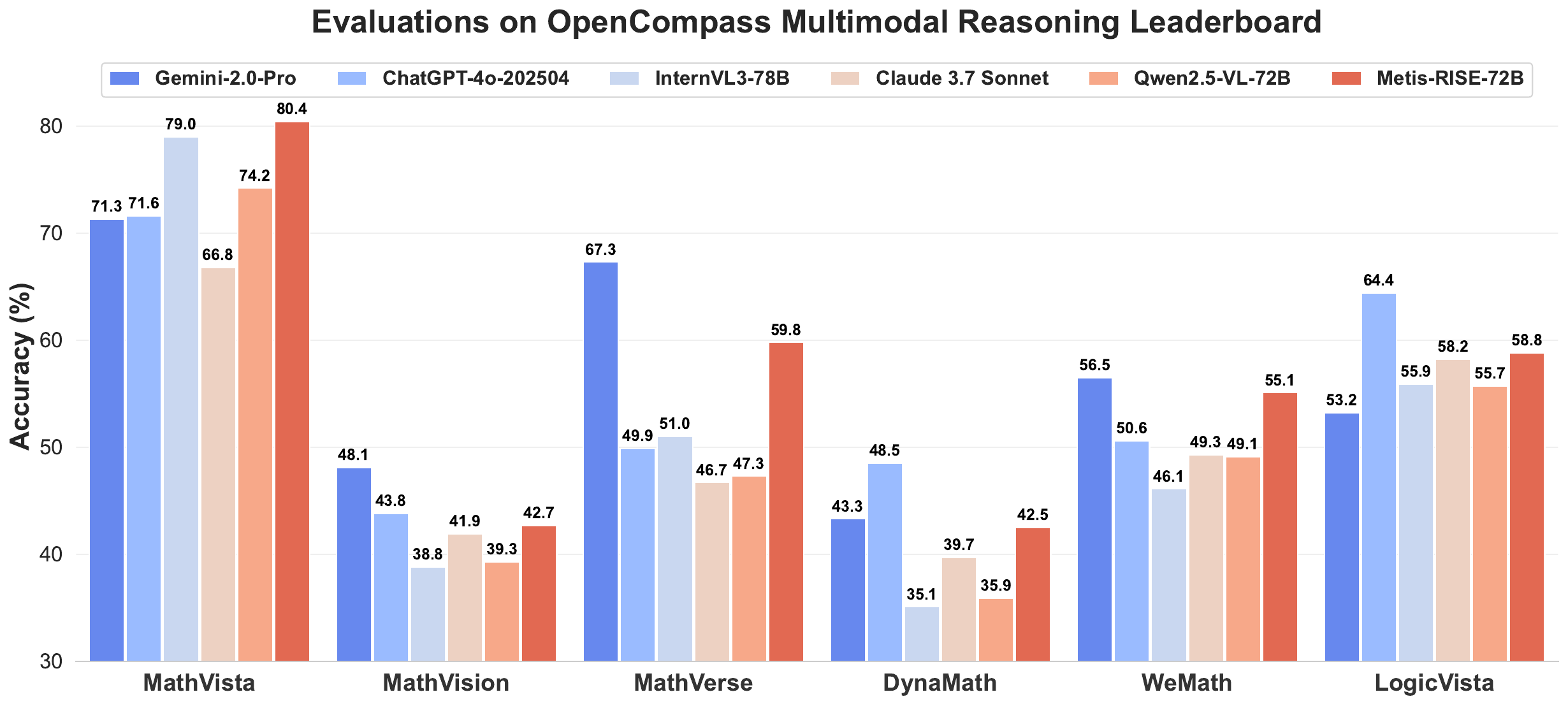}
\caption{We benchmark the proposed \model on the OpenCompass Multimodal Reasoning Leaderboard, comparing it with other state-of-the-art methods.}
\label{fig:comparisons_sota}
\end{figure}

Recently, the field of Large Language Models (LLMs) has witnessed a remarkable surge in the development of reasoning models, garnering substantial attention from the research community. Notable advancements include pioneering works such as DeepSeek-R1~\citep{deepseek-r1}, OpenAI-O1~\citep{openai-o1}, and Seed1.5-Thinking~\citep{seed2025seed}. These models primarily leverage large-scale Reinforcement Learning (RL) training to enhance the reasoning capabilities of their foundation models, thereby enabling them to tackle complex scenarios effectively.

This progress has also catalyzed a flurry of activity in the multimodal domain, with researchers beginning to explore similar paradigms to improve the reasoning capabilities of multimodal large language models (MLLMs)~\citep{peng2025lmmr1,yang2025r1onevision,shen2025vlm,chen2025r1v,meng2025mmeureka,wang2025vl}. The goal is to equip models with the ability to fully comprehend visual signals and perform accurate reasoning in complex multimodal contexts. However, these methods often fall short in several critical aspects. Some studies have only explored models of relatively small scales, such as 3B or 7B parameters, without scaling up to larger models. This limitation undermines the robustness and generalizability of their conclusions. Others focus solely on the RL training phase to boost reasoning performance, neglecting the inherent weaknesses and potential ceilings of foundation models. This leads to two key issues: low sample efficiency during exploration where the success rate across $K$ attempts ranges between 0 and 1 and the inability to activate reasoning capabilities that are entirely absent in the foundation model.

\begin{figure}[t]
\centering
\includegraphics[width=1.0\linewidth]{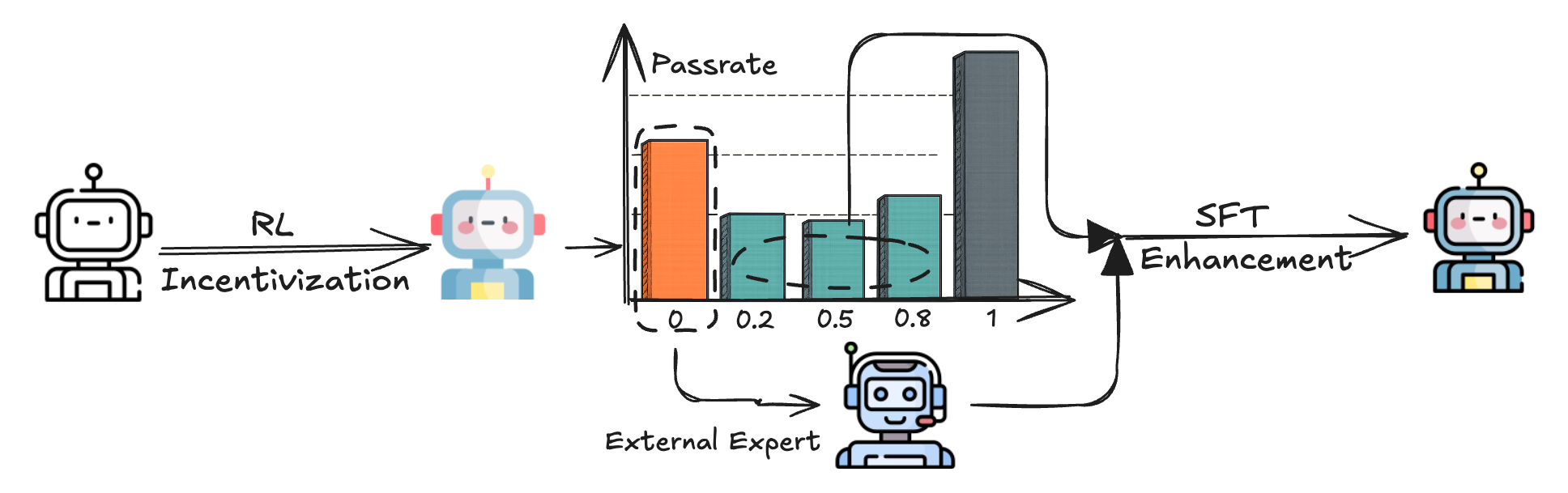}
\caption{Overview of the \model framework. RL first incentivizes exploration and activates latent reasoning. Subsequent SFT stages enhance these abilities by addressing inefficient trajectory sampling (via self-distillation) and fundamental capability absence (via expert-augmented knowledge injection).}
\label{fig:framework}
\end{figure}

To address these limitations, we introduce \textbf{\model}, as illustrated in Figure~\ref{fig:framework}, a hybrid training framework that combines RL incentivization and SFT enhancement in a non-traditional sequence to more effectively boost the reasoning capabilities of multimodal large language models (MLLMs). Unlike conventional pipelines that begin with a cold-start SFT phase, our approach omits this initial step based on empirical observations indicating that early SFT can restrict the model’s exploratory capacity during subsequent RL training, a phenomenon also noted in studies such as VLAA-Thinking~\citep{chen2025sftrlearlyinvestigation} and LLama4~\citep{llama4}. Instead, \model begins with direct RL training using a variant of Group Relative Policy Optimization (GRPO) algorithm. This initial phase incentivizes the model's reasoning potential, but suffers from two major drawbacks: (1) \textit{inefficient trajectory sampling} for problems where the model possesses but struggles to consistently apply the correct reasoning, and (2) \textit{fundamental capability absence} where the foundation model completely fails to yield correct trajectory.

To mitigate these challenges, the SFT phase in \model strategically enhances the model using a curated dataset composed of two parts, directly targeting the aforementioned issues:
\begin{itemize}[left=10pt]
    \item \textbf{Self-Distilled Reasoning Trajectories (to address inefficient sampling):} We perform $K$-shot sampling on the prompt pool using the RL-trained model. For prompts where the model demonstrates inconsistent success (trajectory correctness scores strictly between 0 and 1), we use the model's own correct reasoning trajectories as supervision signals. This reinforces effective reasoning paths the model can discover but does not yet execute reliably.
    \item \textbf{Expert-Augmented Knowledge Injection (to address capability absence):} For prompts where the model consistently fails (trajectory correctness score is 0 across all attempts), we infer a fundamental lack of the necessary reasoning capability. In these cases, a stronger external reasoning expert generates high-quality trajectories. These expert-generated solutions are then used to augment the SFT dataset, effectively injecting new knowledge and compensating for the model's original shortcomings.
\end{itemize}

This strategic application of RL for incentivization followed by SFT for enhancement forms the core of \model, which we leveraged to train two versions MLLM with 7B and 72B parameters. Evaluations on the OpenCompass Multimodal Reasoning Leaderboard, which comprises multiple multimodal mathematical reasoning benchmarks including MathVista~\citep{mathvista}, MathVision~\citep{mathvision}, MathVerse~\citep{mathverse}, DynaMath~\citep{dynamath}, WeMath~\citep{qiao2024we}, and LogicVista~\citep{xiao2024logicvista}, demonstrate the effectiveness of our approach. Both models achieve state-of-the-art performance among similar-sized models, with the 72B version ranking fourth overall on the full leaderboard, validating the scalability and efficacy of \model.

\section{Method}
\label{section:method}

\subsection{RL Incentivization}
\label{subsection:rl}

Our approach is inspired by recent advancements in reasoning-enhanced LLMs, particularly the RL-driven methodologies employed in works such as DeepSeek-R1~\citep{deepseek-r1} and Seed1.5-Thinking~\citep{seed2025seed}. We adopt Group Relative Policy Optimization (GRPO)~\citep{shao2024deepseekmath} as our core RL algorithm due to its effectiveness in aligning model behavior with human preferences while reducing computational overhead compared to traditional Proximal Policy Optimization (PPO)~\citep{ppo}. Meanwhile, we further enhance GRPO by empirically transferring and integrating insights from reasoning LLMs such as DAPO~\citep{dapo} and VAPO~\citep{yue2025vapo}, which make the RL training more stable and effective. Below, we first revisit the GRPO framework and then outline our incorporated modifications.

\subsubsection{Group Relative Policy Optimization (GRPO)}
GRPO aims to estimate the advantages of model-generated responses by comparing a group of candidate outputs generated from the same query. Mathematically, given a query-answer pair $(q, a)$, the behavior policy model $\pi_{\theta_\text{old}}$ generates a group of $G$ candidate trajectories $\{ \tau_i \}_{i=1}^G$. For $i$-th trajectory, the advantage at generation time step $t$ is computed by normalizing the group-level rewards $\{ R_i \}_{i=1}^G$ as follows:
\begin{equation}
\hat{A}_{i,t} = \frac{R_i - \text{mean}(\{R_i\}_{i=1}^G)}{\text{std}(\{R_i\}_{i=1}^G)}.
\end{equation}

The GRPO objective integrates a clipped surrogate objective along with a KL-regularization term that constrains the deviation between the current policy $\pi_\theta$ and a reference policy $\pi_\text{ref}$:
\begin{equation}
\begin{aligned}
& \mathcal{J}_\text{GRPO}(\theta) = \mathbb{E}_{(q,a)\sim \mathcal{D}, \{\tau_i\}_{i=1}^G\sim \pi_{\theta_\text{old}}(\cdot\mid q)} \\
&\ \Bigg[ \frac{1}{G}\sum_{i=1}^{G} \frac{1}{|\tau_i|}\sum_{t=1}^{|\tau_i|} \Bigg( 
\min \Big( r_{i,t}(\theta) \hat{A}_{i,t},  
\ \text{clip} \Big( r_{i,t}(\theta), 1 - \varepsilon, 1 + \varepsilon \Big) \hat{A}_{i,t} \Big)
- \beta D_{\text{KL}}(\pi_{\theta} || \pi_{\text{ref}}) 
\Bigg) \Bigg],
\end{aligned}
\label{eq:grpo_ori}
\end{equation}
where $r_{i,t}(\theta)$ denotes the importance sampling ratio at time step $t$ of the $i$-th response, defined as:
\begin{equation}
r_{i,t}(\theta) = \frac{\pi_{\theta}(\tau_{i,t} \mid q, \tau_{i,<t})}{\pi_{\theta_{\text{old}}}(\tau_{i,t} \mid q, \tau_{i,<t})}.
\end{equation}

In this formulation, $\varepsilon > 0$ controls the tolerance for policy updates, ensuring stable training by limiting large deviations in the importance weights. The hyperparameter $\beta$ governs the strength of the regularization that penalizes divergence from the reference policy.

\subsubsection{Incorporated Modifications}
To enhance the stability and effectiveness of the GRPO training process, we empirically draw on advances in reasoning-oriented LLM~\citep{dapo,yue2025vapo} and adapt these insights to multimodal learning, including elimination of KL divergence, online data filtering, asymmetric coupling, token-level policy loss, and soft overlong punishment. Detailed explanations of these modifications are provided in Appendix~\ref{app:rl_modifications}. Based on the original GRPO defined by Equation~\ref{eq:grpo_ori}, we incorporate the aforementioned modifications (highlighted with blue color) and lead to the final formulation as follows:

\begin{equation}
\begin{aligned}
\mathcal{J}_{\text{DAPO}}(\theta) =\quad& \mathbb{E}_{(q,a)\sim \mathcal{D}, \{\tau_i\}_{i=1}^G\sim \pi_{\theta_\text{old}}(\cdot\mid q)}\\&
\Bigg[\frac{1}{\color{citeblue}\sum_{i=1}^{G}|\tau_i|}{\color{citeblue}\sum_{i=1}^{G}\sum_{t=1}^{|\tau_i|}} 
\min \Big( r_{i,t}(\theta) \hat{A}_{i,t},  
\ \text{clip} \Big( r_{i,t}(\theta), 1 - {\color{citeblue}\varepsilon_{\text{low}}}, 1 + {\color{citeblue}\varepsilon_{\text{high}}} \Big) \hat{A}_{i,t} \Big) \Bigg]
\\
\text{s.t.}\quad& {\color{citeblue}0< \Big|\{\tau_i\mid\texttt{is\_equal}(gt,\tau_i)\}\Big|< G}.
\label{eq:dapoloss_oversample_filter}
\end{aligned}
\end{equation}
where
\begin{equation}
    r_{i,t}(\theta)=\frac{\pi_{\theta}(\tau_{i,t} \mid q, \tau_{i,<t})}{\pi_{\theta_{\text{old}}}(\tau_{i,t} \mid q,\tau_{i,<t})},\quad\hat{A}_{i,t} = \frac{R_i - \text{mean}(\{R_i\}_{i=1}^G)}{\text{std}(\{R_i\}_{i=1}^G)}.
\label{eq:advantage_calculation}
\end{equation}

\subsubsection{Reward Design}

Our approach leverages \textbf{verifiable data} during the training phase, which allows the correctness of the model's responses to be evaluated via a rule-based verifier. Following DeepSeek-R1~\citep{deepseek-r1}, we employ a hybrid reward structure that combines both \textbf{format reward} and \textbf{accuracy reward}, as detailed below. 
\begin{itemize}[left=10pt]
    \item \textbf{format reward}: it ensures that the model generates responses strictly adhering to a predefined format. Specifically, the model must first put its thinking process in \verb|<think>| and \verb|</think>|, and enclose its final answer within the delimiters \verb|<answer>| and \verb|</answer>|. If this syntactic requirement is not met, the format reward is set to zero. This encourages the model to conform to a structured output format, facilitating thinking activation and verification.
    \item \textbf{accuracy reward}: it evaluates the semantic correctness of the generated response. Once the model produces an answer within the designated format, the system extracts the content inside the \verb|<answer>| tags and feeds it into the rule-based verifier. If the verifier confirms the correctness of the answer, the accuracy reward is assigned a value of $1$; otherwise, it is set to $0$. This binary signal provides clear supervision for the model to learn accurate reasoning and response generation.
\end{itemize}

During training, the following system prompt is used to guide the model’s behavior:
\begin{promptbox}{System Prompt}
\begin{verbatim}
Solve the question. The user asks a question, and you solves it. You first 
thinks about the reasoning process in the mind and then provides the user with 
the answer. The answer is in latex format and wrapped in $...$. The final answer 
must be wrapped using the \\boxed{} command. The reasoning process and answer 
are enclosed within <think> </think> and <answer> </answer> tags, respectively, 
i.e., <think> Since $1+1=2$, so the answer is $2$. </think><answer> The answer 
is $\\boxed{2}$ </answer>, which means assistant's output should start with 
<think> and end with </answer>.
\end{verbatim}
\end{promptbox}

\subsection{SFT Enhancement}
\label{subsection:sft}
The initial phase, as described above, effectively incentivizes  the model’s reasoning capabilities through reinforcement learning (RL). However, this approach is subject to two primary limitations:

\begin{itemize}[left=10pt]
    \item \textit{Inefficient Trajectory Sampling:} Although the model may already possess the correct reasoning logic, it often fails to consistently generate valid reasoning paths due to suboptimal exploration or sampling strategies.

    \item \textit{Fundamental Capability Absence:} In certain cases, the model entirely lacks the requisite reasoning ability to solve specific types of problems, resulting in persistent failure even after multiple sampling attempts.
\end{itemize}

To address these shortcomings, we introduce a subsequent supervised fine-tuning (SFT) stage designed to further refine the model using a carefully curated dataset comprising two distinct components:

\begin{itemize}[left=10pt]
    \item \textbf{Self-Distilled Reasoning Trajectories:}
    Given a prompt pool $\mathcal{P} = \{p_1, p_2, \ldots, p_N\}$, we perform $K$-shot reasoning trajectory sampling using the RL-trained model $M_{\text{RL}}$. For each prompt $p_i$, we collect a set of reasoning trajectories $\mathcal{T}_i = \{\tau_{i1}, \tau_{i2}, \ldots, \tau_{iK}\}$, where each $\tau_{ik}$ denotes a response generated by the model during problem solving. The correctness of each trajectory is evaluated using a rule-based verifier $v(\tau_{ik}) \in [0, 1]$. Prompts are then categorized based on their overall correctness score $v(p_i)$, calculated as the average over all sampled trajectories. We specifically select prompts for which $0 < v(p_i) < 1$, excluding those with perfect ($v(p_i) = 1$) or completely failed ($v(p_i) = 0$) reasoning. For each selected prompt $p_i$, we randomly sample one correct trajectory $\tau^*_i = \text{random}(\tau \mid \tau \in \mathcal{T}_i, v(\tau) = 1)$ to serve as the supervision signal:
    \begin{equation}
        \mathcal{D}_{\text{self}} = \left\{(p_i, \tau^*_i) \mid 0 < v(p_i) < 1 \right\}.
    \end{equation}
    This procedure reinforces reasoning paths that the model is capable of discovering but currently unable to apply reliably.

    \item \textbf{Expert-Augmented Knowledge Injection:}
    For prompts $p_i$ where $v(p_i) = 0$ despite multiple sampling attempts, we infer that the model fundamentally lacks the necessary reasoning capability. To address this deficiency, we employ a stronger external reasoning expert $E$, such as advanced reasoning proprietary models~\citep{doubao_1_5_pro}, to generate high-quality solutions $\tau^e_i$. These expert-generated reasoning trajectories are then incorporated into the SFT dataset:
    \begin{equation}
    \mathcal{D}_{\text{expert}} = \left\{(p_i, \tau^e_i) \mid v(p_i) = 0 \right\}.
    \end{equation}
    By integrating these expert-labeled examples, we inject external knowledge into the model, thereby compensating for its inherent reasoning limitations.
\end{itemize}

The final SFT training dataset is constructed by combining both subsets:
\begin{equation}
\mathcal{D}_{\text{SFT}} = \mathcal{D}_{\text{self}} \cup \mathcal{D}_{\text{expert}}.
\label{eq:sft_data}
\end{equation}

This hybrid dataset enables the model not only to consolidate its existing reasoning capabilities but also to acquire new ones, leading to significant improvements in both the consistency and coverage of correct reasoning paths.

\section{Experiment}
\label{section:experiment}

\subsection{Implementation and Evaluation}

We build upon the open-source Qwen2.5-VL series~\citep{qwen2_5_vl} and adopt a two-stage training approach combining \textit{RL Incentivization} and \textit{SFT Enhancement} to train two model variants: \model-7B and \model-72B. In the initial RL stage, we train the models on around 40K multimodal reasoning samples by optimizing the objective defined in Equation~\ref{eq:dapoloss_oversample_filter}. This is followed by the SFT enhancement stage, where we curate around 15K training samples according to Equation~\ref{eq:sft_data}, incorporating both self-distilled data $\mathcal{D}_{\text{self}}$ and external expert annotations $\mathcal{D}_{\text{expert}}$.

To comprehensively evaluate the performance of our models, we employ VLMEvalKit~\citep{duan2024vlmevalkit} and benchmark on the \href{https://rank.opencompass.org.cn/leaderboard-multimodal-reasoning/?m=REALTIME}{OpenCompass Multimodal Reasoning Leaderboard}, which aggregates multiple challenging benchmarks to evaluate multimodal mathematical and logical reasoning capabilities. The evaluation set includes the following six key benchmarks:

\begin{itemize}[left=10pt]
    \item \textbf{MathVista}~\citep{mathvista}: Its mini test split (MathVista\_MINI, around 1,000 samples) for complex mathematical problem-solving using visual and textual inputs.
    \item \textbf{MathVision}~\citep{mathvision}: The full test set (around 3,000 samples) evaluating vision-language models on math tasks requiring visual understanding.
    \item \textbf{MathVerse}~\citep{mathverse}: Its mini test split (MathVerse\_MINI, around 700 samples) in "Vision Only" mode, focusing on visual cross-modal reasoning.
    \item \textbf{DynaMath}~\citep{dynamath}: The full test set (around 5,000 samples from 501 original questions with 10 variations each) designed for dynamic multi-step mathematical reasoning.
    \item \textbf{WeMath}~\citep{qiao2024we}: Its mini test split (around 1,740 samples) featuring real-world word problems requiring mathematical computation and contextual understanding. We report \textit{Score (Strict)}.
    \item \textbf{LogicVista}~\citep{xiao2024logicvista}: The full test set (around 450 samples) targeting multimodal logical inference capabilities.
\end{itemize}

We compare our \model against three categories of state-of-the-art models:

\begin{itemize}[left=10pt]
    \item \textbf{Proprietary Models}: These include Doubao-1.5-Pro~\citep{doubao_1_5_pro}, Gemini-2.0-Pro~\citep{Gemini2}, ChatGPT-4o-202504~\citep{gpt4o}, Gemini-2.0-Flash~\citep{Gemini2}, Claude 3.7 Sonnet~\citep{Claude37}, and GLM-4v-Plus-20250111~\citep{GLM4VPlus}.
    
    \item \textbf{Open-source $\leq$10B Models}: Including Kimi-VL-A3B-Instruct~\citep{kimiteam2025kimivltechnicalreport}, Qwen2.5-VL-7B~\citep{qwen2_5_vl}, InternVL3-8B~\citep{zhu2025internvl3}, and VLAA-Thinker-7B~\citep{chen2025sftrlearlyinvestigation}.
    
    \item \textbf{Open-source $>$10B Models}: Including InternVL3-14B~\citep{zhu2025internvl3}, Ovis2-34B~\citep{lu2024ovis}, QVQ-72B-Preview~\citep{qvq-72b-preview}, LLaVA-OneVision-72B~\citep{li2024llavaonevisioneasyvisualtask}, Qwen2.5-VL-72B~\citep{qwen2_5_vl}, and InternVL3-78B~\citep{zhu2025internvl3}.
\end{itemize}

\subsection{Quantitative Results}
\subsubsection{Comparisons Against SOTA}
\begin{table*}[ht]
\centering
\resizebox{\textwidth}{!}{
\setlength{\tabcolsep}{3pt}
\begin{tabular}{lccccccc}
\toprule
\textbf{Model} & \textbf{Avg.} & \textbf{MathVista} & \textbf{MathVision} & \textbf{MathVerse} & \textbf{DynaMath} & \textbf{WeMath} & \textbf{LogicVista} \\
\midrule
\rowcolor{gray!20}
\multicolumn{8}{c}{\textit{\textbf{Proprietary Models}}} \\
Doubao-1.5-Pro & 61.6 & 78.6 & 51.5 & 64.7 & 44.9 & 65.7 & 64.2 \\
Gemini-2.0-Pro & 56.6 & 71.3 & 48.1 & 67.3 & 43.3 & 56.5 & 53.2 \\
ChatGPT-4o-202504 & 54.8 & 71.6 & 43.8 & 49.9 & 48.5 & 50.6 & 64.4 \\
Gemini-2.0-Flash & 50.6 & 70.4 & 43.6 & 47.8 & 42.1 & 47.4 & 52.3 \\
Claude 3.7 Sonnet & 50.4 & 66.8 & 41.9 & 46.7 & 39.7 & 49.3 & 58.2 \\
GLM-4v-Plus-202501 & 49.2 & 73.5 & 51.1 & 40.7 & 27.5 & 47.7 & 54.4 \\
\midrule
\rowcolor{gray!20}
\multicolumn{8}{c}{\textit{\textbf{Open-source $\leq$10B Models}}} \\
Kimi-VL-A3B-Instruct & 35.8 & 66.0 & 21.8 & 34.1 & 18.0 & 32.3 & 42.7 \\
Qwen2.5-VL-7B & 40.1 & 68.1 & 25.4 & 41.1 & 21.8 & 36.2 & 47.9 \\
InternVL3-8B & 41.4 & 70.5 & 30.0 & 38.5 & 25.7 & 39.5 & 44.5 \\
VLAA-Thinker-7B & 42.5 & 68.0 & 26.4 & 48.2 & 22.4 & 41.5 & 48.5 \\
\rowcolor{citeblue!20}
\model-7B & 46.4 & 75.8 & 28.7 & 51.0 & 27.7 & 45.2 & 49.7 \\
\midrule
\rowcolor{gray!20}
\multicolumn{8}{c}{\textit{\textbf{Open-source $>$10B Models}}} \\
InternVL3-14B & 46.0 & 74.4 & 34.0 & 43.7 & 30.3 & 41.3 & 52.1 \\
Ovis2-34B & 47.9 & 76.1 & 31.9 & 50.1 & 27.5 & 51.9 & 49.9 \\
QVQ-72B-Preview & 46.9 & 70.3 & 34.9 & 48.2 & 30.7 & 39.0 & 58.2 \\
LLaVA-OneVision-72B & 34.7 & 67.1 & 25.3 & 27.2 & 15.6 & 32 & 40.9 \\
Qwen2.5-VL-72B & 50.3 & 74.2 & 39.3 & 47.3 & 35.9 & 49.1 & 55.7 \\
InternVL3-78B & 51.0 & 79.0 & 38.8 & 51.0 & 35.1 & 46.1 & 55.9 \\
\rowcolor{citeblue!20}
\model-72B & 56.6 & 80.4 & 42.7 & 59.8 & 42.5 & 55.1 & 58.8 \\
\bottomrule
\end{tabular}}
\caption{Performance comparison of our \model models (\model-7B and \model-72B) against prominent proprietary and other existing models across six mathematical reasoning benchmarks. Results are from the OpenCompass Multimodal Reasoning Leaderboard, detailing both average scores and performance on individual benchmarks.}
\label{tab:main_results}
\end{table*}

\begin{table*}[ht]
\centering
\resizebox{\textwidth}{!}{
\setlength{\tabcolsep}{3pt}
\begin{tabular}{llcccccc}
\toprule
\textbf{Training Recipe} & \textbf{Avg.} & \textbf{MathVista} & \textbf{MathVision} & \textbf{MathVerse} & \textbf{DynaMath} & \textbf{WeMath} & \textbf{LogicVista} \\
\midrule
Baseline$^\dagger$ & 39.2 & 67.4 & 26.2 & 41.1 & 20.2 & 34.5 & 45.6 \\

\  $\hookrightarrow$ RL & 44.0\textcolor{citeblue!80}{\small $_{\mathbf{\uparrow}4.8}$} & 72.8 & 28.7 & 46.8 & 26.2 & 43.3 & 46.5 \\
\qquad $\hookrightarrow$ MM SFT & 45.7\textcolor{citeblue!80}{\small $_{\mathbf{\uparrow}6.5}$} & 75.1 & 29.5 & 47.0 & 26.8 & 45.1 & 51.0 \\ %
\qquad $\hookrightarrow$ Text SFT & 45.5\textcolor{citeblue!80}{\small $_{\mathbf{\uparrow}6.3}$} & 75.6 & 27.3 & 48.7 & 27.9 & 45.4 & 47.9 \\
\qquad $\hookrightarrow$ Mixed SFT & 46.4\textcolor{citeblue!80}{\small $_{\mathbf{\uparrow}7.2}$} & 75.8 & 28.7 & 51.0 & 27.7 & 45.2 & 49.7 \\ %
\bottomrule
\end{tabular}}
\caption{Ablation study results for different training stages and SFT variants (Mixed SFT = MM + Text SFT). Qwen2.5-VL-7B serves as the baseline. $^\dagger$Denotes our own evaluation; official results for this baseline can be found in Table~\ref{tab:main_results}.}
\label{tab:7b_ablation}
\end{table*}

Table \ref{tab:main_results} presents a comprehensive quantitative comparison of our \model models against a range of leading proprietary and other existing models across six challenging mathematical reasoning benchmarks, sourced from the OpenCompass Multimodal Reasoning Leaderboard. Our \model-7B model demonstrates exceptional performance within the $\leq$10B parameter category, achieving an average score of 46.4. This result surpasses all other models of comparable size, including VLAA-Thinker-7B (42.5) and InternVL3-8B (41.4), establishing \model-7B as the state-of-the-art for its scale on these benchmarks.

Further showcasing the strength of our approach, \model-72B achieves an impressive average score of 56.6, positioning it as the top-performing model in the $>$10B parameter category. It significantly outperforms other large models such as InternVL3-78B (51.0) and Qwen2.5-VL-72B (50.3). Notably, the performance of \model-72B is highly competitive with, and in some cases exceeds, that of prominent proprietary models. For instance, \model-72B surpasses ChatGPT-4o-202504 (54.8) and Claude 3.7 Sonnet (50.4), while being comparable to the performance of Gemini-2.0-Pro (56.6). Collectively, these strong results contribute to \model-72B ranking fourth overall on the OpenCompass Multimodal Reasoning Leaderboard at the time of this evaluation, underscoring its advanced capabilities in complex multimodal reasoning tasks.

\subsubsection{Ablation Study}

Table~\ref{tab:7b_ablation} presents a detailed ablation study conducted on our \model-7B, meticulously illustrating the distinct impact and synergistic contributions of each stage within the \model framework. The baseline model (Qwen2.5-VL-7B) establishes an initial average score of 39.2 across the evaluated datasets. Upon applying the initial RL phase (Baseline$\rightarrow$RL), a significant performance uplift is observed, with the average score increasing to 44.0 (+4.8 points). This substantial gain underscores RL's critical role in incentivizing the model's exploratory capacity, encouraging it to discover and activate latent or inconsistently applied reasoning pathways. This effect is particularly evident in challenging datasets such as WeMath, where the score jumps from 36.2 to 43.3, and DynaMath, which sees an improvement from 21.8 to 26.2, demonstrating RL's ability to unlock reasoning potential. 

Subsequently, the targeted SFT stage is employed to further enhance the model's reasoning capabilities. In this phase, we investigate the impact of SFT using data from two distinct modalities: purely textual data (Text SFT) and multimodal image-text datasets (MM SFT). Both SFT datasets are meticulously constructed, incorporating self-distilled reasoning trajectories and trajectories augmented with external expert knowledge, as formulated in Equation~\ref{eq:sft_data}. Furthermore, we explore a Mixed SFT approach by combining data from both modalities. Note that for the SFT dataset, all true/false questions were filtered out. Regarding multiple-choice questions, we removed those with a pass rate below 0.5, while retaining those with a pass rate of 0.5 or higher (up to 1.0). All free-form questions were retained. This filtering was necessary because we observed that multiple-choice and true/false questions are prone to reward hacking, where the model produces incorrect intermediate reasoning steps but arrives at the correct final answer through guessing.

As presented in Table~\ref{tab:7b_ablation}, our ablation studies demonstrate that all SFT variants yield performance gains over the RL-enhanced baseline (which scored $44.0$ on average). Specifically, when applied after the RL stage, MM SFT further improves the average score by $1.7$ points (from $44.0$ to $45.7$), and Text SFT leads to a $1.5$ point increase (to $45.5$); the Mixed SFT approach achieves the best result, boosting the average score by $2.4$ points to $46.4$ over the RL-enhanced model. This additional improvement highlights SFT's effectiveness in refining and consolidating the reasoning abilities that were surfaced and activated by RL. SFT achieves this by addressing specific challenges such as inefficient trajectory sampling during RL or bridging residual knowledge gaps, as exemplified by the notable improvement on the MathVerse benchmark, where Mixed SFT boosts performance by $4.2$ points over the RL-enhanced model (from $46.8$ to $51.0$).

A similar progressive improvement is observed with our larger \model-72B model, where performance consistently increases from a baseline average of $50.3$, to $55.2$ after RL, and further to $56.6$ following SFT. This consistent growth across model scales underscores the general effectiveness of our multi-stage training approach. Collectively, these progressive and cumulative gains across the stages strongly validate the core tenets of the \model strategy: RL serves as a powerful initial catalyst to unlock and incentivize the model's reasoning potential, which is then systematically enhanced and stabilized by the subsequent SFT phase, leading to superior overall multimodal reasoning capabilities.

\subsection{Qualitative Analysis}
\begin{figure}[htbp]
    \centering
    \begin{minipage}{\textwidth} %
        \centering
        \begin{subfigure}[b]{0.48\textwidth}
            \centering
            \includegraphics[width=\textwidth]{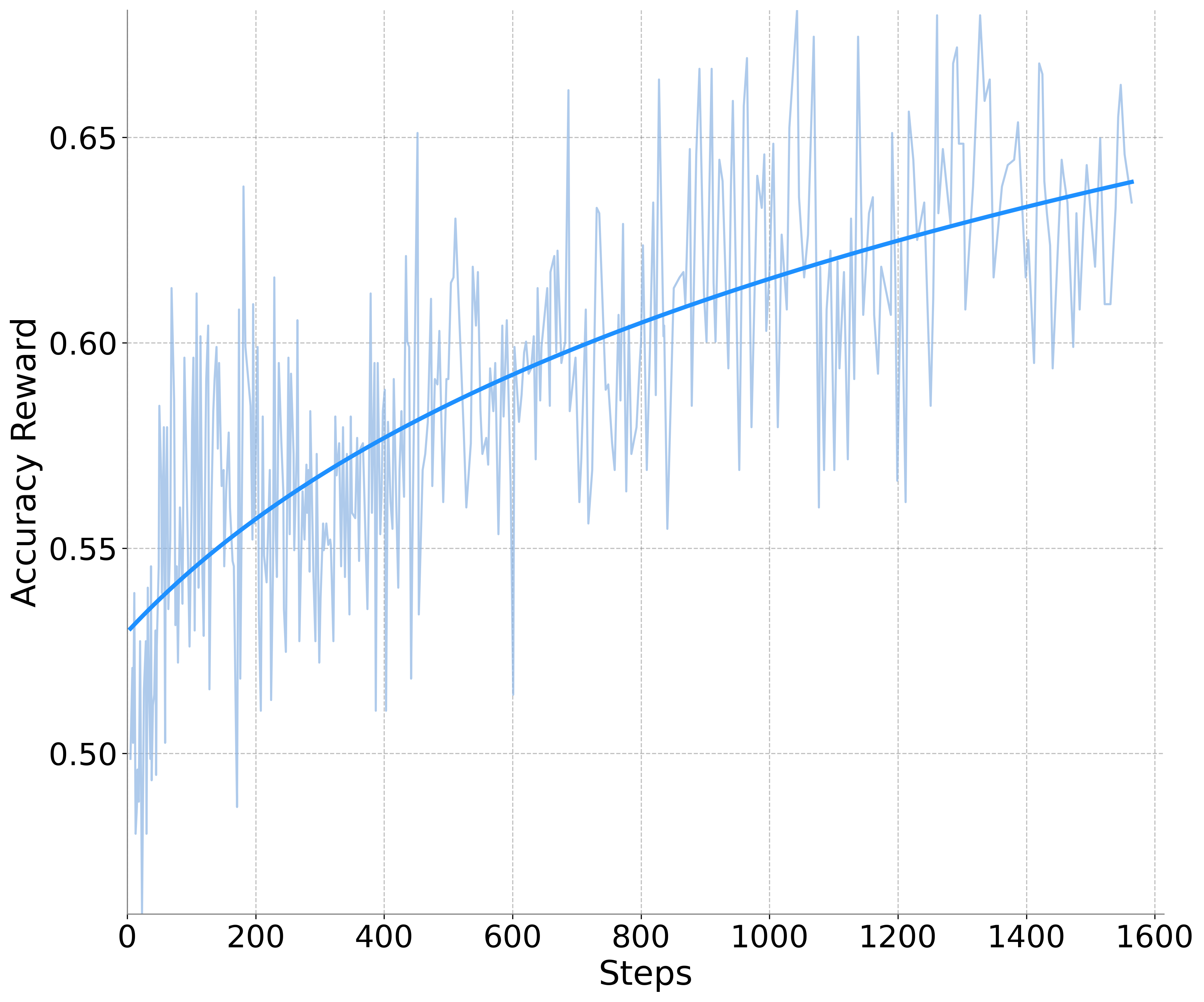}
            \caption{Accuracy Reward over training steps}
            \label{fig:acc_reward_curve}
        \end{subfigure}
        \hfill
        \begin{subfigure}[b]{0.48\textwidth}
            \centering
            \includegraphics[width=\textwidth]{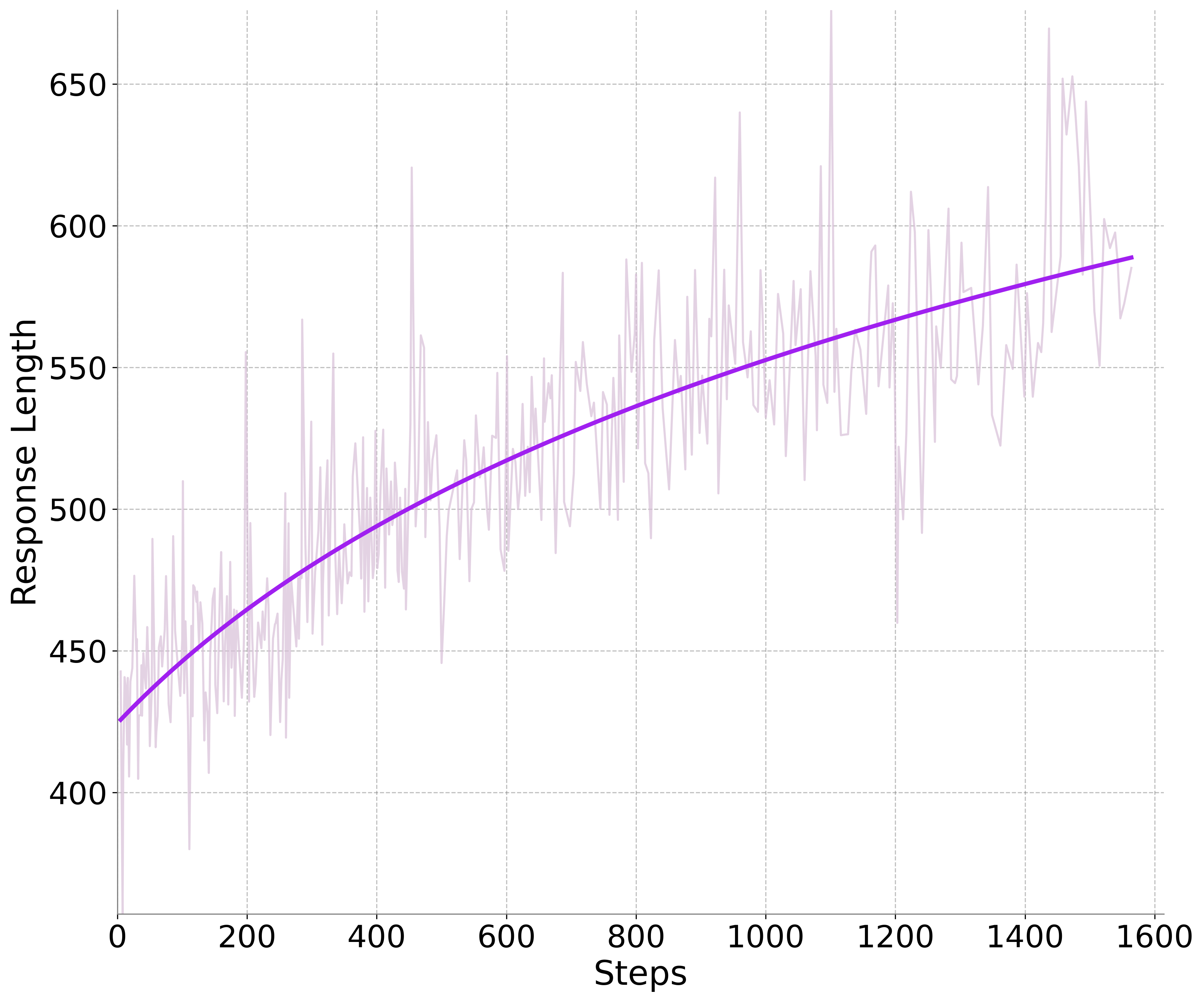}
            \caption{Response Length over training steps}
            \label{fig:response_length_curve}
        \end{subfigure}
    \end{minipage}
    \caption{Training dynamics of Accuracy Reward and Response Length during the initial RL phase for \model-72B. Subfigure (a) shows the progression of accuracy reward, and subfigure (b) illustrates the change in average response length as training proceeds.}
    \label{fig:rl_phase_curves}
\end{figure}

During the initial Reinforcement Learning (RL) phase of our \model-72B model, we observe compelling trends in both accuracy reward and response length, as depicted in Figure~\ref{fig:rl_phase_curves}. Specifically, Figure~\ref{fig:acc_reward_curve} demonstrates a consistent and continuous increase in the accuracy reward as training progresses. Simultaneously, Figure~\ref{fig:response_length_curve} reveals a corresponding upward trend in the average length of the model's responses. This concurrent growth strongly suggests that our model, through RL, learns to engage in more extensive and elaborate reasoning processes. The increasing response length indicates the development of more detailed "chain-of-thought" or step-by-step reasoning, which in turn enables the model to successfully tackle a greater number of problems, thereby leading to the observed rise in accuracy reward. This phenomenon, where an extension in the model's generated thought process correlates with improved problem-solving capabilities, aligns with the trends reported for models like Deepseek-R1~\citep{deepseek-r1} during their respective RL training phases.

\begin{figure}[!ht]
    \centering
    \begin{vqaexample}[width=\textwidth]{Analytic Geometry}
        \begin{minipage}[t]{0.3\textwidth}
            \centering
            \includegraphics[width=\textwidth]{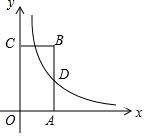}
        \end{minipage}
        \hfill
        \begin{minipage}[t]{0.65\textwidth}
            \begin{tcolorbox}[
                enhanced,
                colback=questionbg,
                colframe=bordercolor!70,
                boxrule=0.8pt,
                arc=3pt,
                left=6pt,
                right=6pt,
                top=4pt,
                bottom=4pt,
                fonttitle=\bfseries\color{titlecolor},
                title=Question: ,
                attach title to upper
            ]
                \ As shown in the figure, the graph of the inverse proportion function $y = \frac{k}{x}$ $(x > 0)$ passes through the midpoint $D$ of side $AB$ of rectangle $OABC$. If the area of rectangle $OABC$ is 8, then $k =$ \_\_\_.
            \end{tcolorbox}
        \end{minipage}
        
        \vspace{0.8em}
        
        \begin{tcolorbox}[
            enhanced,
            breakable,
            colback=answerbg,
            colframe=bordercolor!70,
            boxrule=0.8pt,
            arc=3pt,
            left=6pt,
            right=6pt,
            top=4pt,
            bottom=4pt,
            fonttitle=\bfseries\color{titlecolor},
            title=Answer: ,
            attach title to upper
        ]
            \ \colorbox{gray!10}{\texttt{<think>}}To solve the problem, we need to determine the value of $k$ for the inverse proportion function $y = \frac{k}{x}$ given that it passes through the midpoint $D$ of the side $AB$ of the rectangle $OABC$ and the area of the rectangle is 8.
            
            \begin{enumerate}[leftmargin=*,itemsep=0.2em]
                \item \textbf{Identify the coordinates of the points:}
                \begin{itemize}[leftmargin=*,itemsep=0.1em]
                    \item Let the coordinates of $O$ be $(0, 0)$.
                    \item Let the coordinates of $A$ be $(a, 0)$.
                    \item Let the coordinates of $C$ be $(0, b)$.
                    \item Therefore, the coordinates of $B$ will be $(a, b)$.
                \end{itemize}
                
                \item \textbf{Determine the midpoint $D$ of $AB$:}
                \begin{itemize}[leftmargin=*,itemsep=0.1em]
                    \item The coordinates of $D$ are the average of the coordinates of $A$ and $B$.
                    \item So, $D \left( \frac{a + a}{2}, \frac{0 + b}{2} \right) = \left( a, \frac{b}{2} \right)$.
                \end{itemize}
                
                \item \textbf{Use the area of the rectangle to find the relationship between $a$ and $b$:}
                \begin{itemize}[leftmargin=*,itemsep=0.1em]
                    \item The area of rectangle $OABC$ is given by:
                    $$a \times b = 8$$
                \end{itemize}
                
                \item \textbf{Substitute the coordinates of $D$ into the inverse proportion function:}
                \begin{itemize}[leftmargin=*,itemsep=0.1em]
                    \item The function $y = \frac{k}{x}$ passes through $D \left( a, \frac{b}{2} \right)$.
                    \item So, substituting $x = a$ and $y = \frac{b}{2}$ into the function, we get:
                    $$\frac{b}{2} = \frac{k}{a}$$
                \end{itemize}
                
                \item \textbf{Solve for $k$:}
                \begin{itemize}[leftmargin=*,itemsep=0.1em]
                    \item Rearrange the equation to solve for $k$:
                    $$k = a \times \frac{b}{2} = \frac{ab}{2}$$
                    \item Since $ab = 8$ (from the area of the rectangle), we substitute $8$ for $ab$:
                    $$k = \frac{8}{2} = 4$$
                \end{itemize}
            \end{enumerate}
            
            \vspace{0.3em}
            Thus, the value of $k$ is $4$.
            
            The answer is $\boxed{4}$.\colorbox{gray!10}{\texttt{</think>}}
            
            \vspace{0.3em}
            \colorbox{gray!10}{\texttt{<answer>}}The answer is $\boxed{4}$\colorbox{gray!10}{\texttt{</answer>}}
        \end{tcolorbox}
    \end{vqaexample}
    \caption{Example of a multi-step analytic geometry problem solved by \model-72B.}
    \label{fig:inverse_proportion_example}
\end{figure}

The provided example (Figure \ref{fig:inverse_proportion_example}) compellingly demonstrates the model's strong mathematical reasoning capabilities in a multi-step geometry and algebra problem. The model's detailed thought process showcases a systematic approach, beginning with the correct geometric interpretation and assignment of coordinates to the rectangle's vertices. It then accurately applies the midpoint formula to determine the coordinates of point $D$ and successfully translates the given area constraint into the algebraic expression $ab=8$. Furthermore, the model correctly utilizes the properties of the inverse proportion function by substituting the coordinates of $D$ into $y=k/x$. Crucially, it integrates these pieces of information through adept algebraic manipulation, combining the derived equations to accurately solve for the unknown constant $k$. This ability to not only recall and apply individual mathematical concepts but also to weave them into a coherent, step-by-step reasoning chain to reach the correct solution underscores its advanced problem-solving proficiency. More examples can be found in the Appendix~\ref{app:case}.

\newpage

\newpage

\section{Conclusion}
\label{section:conclusion}

In this work, we introduced \model, a hybrid training paradigm that strategically sequences RL and SFT to significantly advance multimodal reasoning in MLLMs. By prioritizing RL-driven exploration, \model incentivizes the model to unlock latent reasoning skills and avoids premature convergence often seen in SFT-first approaches. Subsequently, targeted SFT stages enhance these capabilities by efficiently addressing inconsistent reasoning through self-distilled trajectories and rectifying fundamental capability absence via expert knowledge injection. Our evaluations on the OpenCompass Multimodal Reasoning Leaderboard demonstrate that MLLMs trained with \model, at both 7B and 72B scales, achieve state-of-the-art performance among comparable size models, underscoring its effectiveness and scalability. This study highlights that \model's distinctive RL-incentivizes, SFT-enhances methodology offers a potent pathway to improving the reasoning capacity of MLLMs.

\section{Future Work}
\label{section:future_work}

Looking ahead, several promising avenues for future research emerge from this work. First, we plan to explore an iterative application of the \model framework. This would involve cyclically applying RL to incentivize further exploration and SFT to enhance newly activated or refined capabilities (i.e., RL $\rightarrow$ SFT $\rightarrow$ RL $\rightarrow$ SFT...). The aim is to investigate whether such a multi-stage, progressive training paradigm can lead to continued improvements in the model's reasoning abilities, potentially pushing the boundaries of its proficiency even further. 

Second, we intend to extend \model to non-verifiable data domains by developing a model-based verifier. Although the current work leverages verifiable data, allowing for rule-based verifiers during RL training (e.g., for mathematical problems), many complex multimodal reasoning tasks lack easily verifiable ground truths. To address this, we will investigate the use of a trained, model-based verifier to assess the quality and correctness of the MLLM's outputs during the RL phase. This advancement would be crucial for broadening the applicability of \model to a wider array of general multimodal reasoning scenarios beyond mathematical or formally verifiable tasks, encompassing areas requiring more nuanced, open-ended, or subjective evaluation.

\vspace{10mm}
\bibliography{neurips_2025}
\bibliographystyle{unsrtnat}

\clearpage

\appendix

\section{Details of GRPO Modifications}
\label{app:rl_modifications}
We empirically leverage advancements in reasoning-oriented large language models~\citep{dapo,yue2025vapo} and integrate these insights into multimodal learning to improve the stability and effectiveness of the GRPO training process in the following ways.

\begin{itemize}[left=10pt]
    \item \textbf{Elimination of KL Divergence}: The KL divergence regularization term is conventionally employed to constrain the discrepancy between the online policy and a frozen reference policy. This mechanism aims to align model behavior closely with the initial model, preventing excessive deviation. However, in the context of training long-CoT reasoning models, it is observed that the distribution of the evolving model can significantly diverge from its initial state. Consequently, enforcing this constraint may not only become unnecessary but could also impede the model's exploration capabilities, limiting its ability to discover novel solutions. We thus opt to eliminate the KL divergence term.
    
    \item \textbf{Online Data Filtering}: 
    Leveraging the inherent group-wise relative learning mechanism, we implement an online data filtering strategy that dynamically excludes prompts for which all $K$-shot sampled trajectories result in either uniform failure or success. Only those instances yielding accuracy scores strictly within the interval $(0,1)$ and thus producing meaningful gradient signals are retained. This approach helps mitigate gradient signal attenuation during training, promoting more stable and effective model learning.

    \item \textbf{Asymmetric Clipping}: 
    We employ an asymmetric clipping mechanism by decoupling the conventional single clip parameter $\varepsilon$ into distinct lower and upper bounds, denoted as $\varepsilon_\text{low}$ and $\varepsilon_\text{high}$, respectively. By increasing $\varepsilon_\text{high}$, we allow greater flexibility for increases in low-probability tokens while maintaining tighter control over decreases. This encourages broader exploration of the response space, thereby enhancing the model’s capacity to discover novel and effective solutions.

    \item \textbf{Token-Level Policy Loss}: 
    Rather than defining the policy loss at the level of entire responses, we refine it to operate at the token level by aggregating over all individual tokens. This fine-grained formulation addresses potential imbalances in token-level contributions to the overall loss, ensuring that each token's influence on the gradient update is appropriately weighted.

    \item \textbf{Soft Overlong Punishment}:
    For responses exceeding a predefined maximum length, we define a punishment interval wherein the penalty increases proportionally with the excess length. 
\end{itemize}

\newpage

\section{Case Study}
\label{app:case}

\begin{figure}[!ht]
    \centering
    \begin{vqaexample}[width=\textwidth]{Plane Geometry}
        \begin{minipage}[t]{0.25\textwidth}
            \centering
            \includegraphics[width=\textwidth]{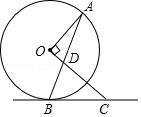}
        \end{minipage}
        \hfill
        \begin{minipage}[t]{0.65\textwidth}
            \begin{tcolorbox}[
                enhanced,
                colback=questionbg,
                colframe=bordercolor!70,
                boxrule=0.8pt,
                arc=3pt,
                left=6pt,
                right=6pt,
                top=4pt,
                bottom=4pt,
                fonttitle=\bfseries\color{titlecolor},
                title=Question: ,
                attach title to upper
            ]
                \ In the given diagram, AB is a chord of circle O. Line OC is perpendicular to OA and intersects tangent BC at point C, intersecting AB at point D. If angle OAB is 20°, what is the degree measure of angle OCB?
            \end{tcolorbox}
        \end{minipage}
        
        \vspace{0.8em}
        
        \begin{tcolorbox}[
            enhanced,
            breakable,
            colback=answerbg,
            colframe=bordercolor!70,
            boxrule=0.8pt,
            arc=3pt,
            left=6pt,
            right=6pt,
            top=4pt,
            bottom=4pt,
            fonttitle=\bfseries\color{titlecolor},
            title=Answer: ,
            attach title to upper
        ]
            \ \colorbox{gray!10}{\texttt{<think>}}\\
            To solve for the degree measure of angle $\angle OCB$, we will analyze the given geometric configuration step by step.
            
            \begin{enumerate}[leftmargin=*,itemsep=0.2em]
                \item \textbf{Identify the given information:}
                \begin{itemize}[leftmargin=*,itemsep=0.1em]
                    \item $AB$ is a chord of circle $O$.
                    \item $OC$ is perpendicular to $OA$ and intersects the tangent $BC$ at point $C$.
                    \item $\angle OAB = 20^\circ$.
                \end{itemize}
                
                \item \textbf{Use the properties of the circle and the tangent:}
                \begin{itemize}[leftmargin=*,itemsep=0.1em]
                    \item Since $OC$ is perpendicular to $OA$, $\angle AOC = 90^\circ$.
                    \item $BC$ is a tangent to the circle at point $B$, so $\angle OBC = 90^\circ$ (angle between a radius and a tangent at the point of tangency).
                \end{itemize}
                
                \item \textbf{Determine the angle $\angle AOB$:}
                \begin{itemize}[leftmargin=*,itemsep=0.1em]
                    \item In $\triangle OAB$, $OA = OB$ (radii of the circle), so $\triangle OAB$ is isosceles.
                    \item The sum of angles in a triangle is $180^\circ$.
                    \item $\angle OAB = \angle OBA = 20^\circ$.
                    \item Therefore, $\angle AOB = 180^\circ - 20^\circ - 20^\circ = 140^\circ$.
                \end{itemize}
                
                \item \textbf{Find $\angle BOC$:}
                \begin{itemize}[leftmargin=*,itemsep=0.1em]
                    \item Since $\angle AOC = 90^\circ$ and $\angle AOB = 140^\circ$, $\angle BOC = \angle AOB - \angle AOC = 140^\circ - 90^\circ = 50^\circ$.
                \end{itemize}
                
                \item \textbf{Determine $\angle OCB$:}
                \begin{itemize}[leftmargin=*,itemsep=0.1em]
                    \item In $\triangle OBC$, $\angle OBC = 90^\circ$.
                    \item The sum of angles in a triangle is $180^\circ$.
                    \item $\angle OCB = 180^\circ - \angle OBC - \angle BOC = 180^\circ - 90^\circ - 50^\circ = 40^\circ$.
                \end{itemize}
            \end{enumerate}
            
            \vspace{0.3em}
            Thus, the degree measure of $\angle OCB$ is $40^\circ$.
            
            The final answer is:\\
            \colorbox{gray!10}{\texttt{</think>}}
            
            \colorbox{gray!10}{\texttt{<answer>}}\\
            The answer is $\boxed{40}$\\
            \colorbox{gray!10}{\texttt{</answer>}}
        \end{tcolorbox}
    \end{vqaexample}
    \label{fig:example_geo}
\end{figure}

\newpage

\begin{figure}[!ht]
    \centering
    \begin{vqaexample}[width=\textwidth]{Chart Reasoning}
        \begin{minipage}[t]{0.45\textwidth}
            \centering
            \includegraphics[width=\textwidth]{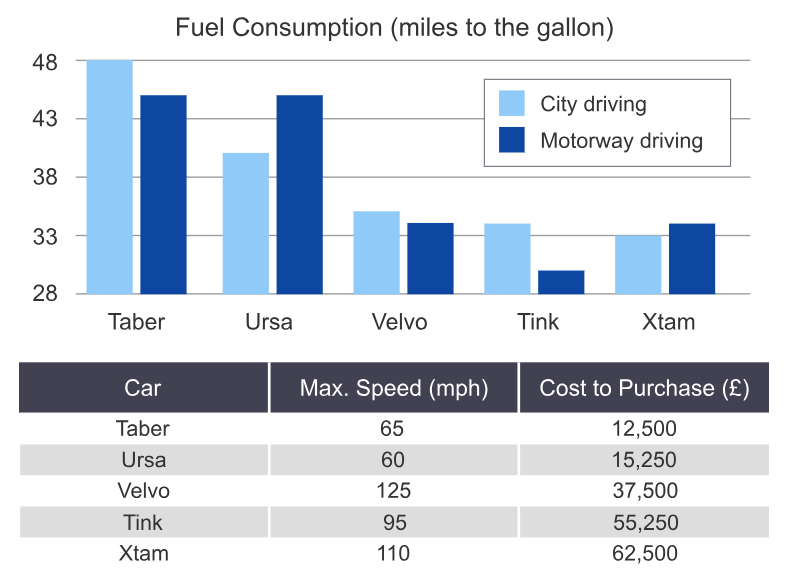}
        \end{minipage}
        \hfill
        \begin{minipage}[t]{0.52\textwidth}
            \begin{tcolorbox}[
                enhanced,
                colback=questionbg,
                colframe=bordercolor!70,
                boxrule=0.8pt,
                arc=3pt,
                left=6pt,
                right=6pt,
                top=4pt,
                bottom=4pt,
                fonttitle=\bfseries\color{titlecolor},
                title=Question: ,
                attach title to upper
            ]
                \ If a driver travels an average of 4,250 miles per month driving only along motorways in an Xtam car, what is the predicted annual consumption of fuel (in gallons)? Select from A, B, C, D and E. (A)Cannot say (B) 1,500 (C) 125 (D) 150 (E) 1,250
            \end{tcolorbox}
        \end{minipage}
        
        \vspace{0.8em}
        
        \begin{tcolorbox}[
            enhanced,
            breakable,
            colback=answerbg,
            colframe=bordercolor!70,
            boxrule=0.8pt,
            arc=3pt,
            left=6pt,
            right=6pt,
            top=4pt,
            bottom=4pt,
            fonttitle=\bfseries\color{titlecolor},
            title=Answer: ,
            attach title to upper
        ]
            \ \colorbox{gray!10}{\texttt{<think>}}\\
            To solve this problem, we need to follow these steps:

        1. \textbf{Identify the fuel consumption of the Xtam car for motorway driving:}
        
        \begin{itemize}[left=10pt,itemsep=0.1em]
        \item From the bar chart, the fuel consumption for Xtam in motorway driving is 33 miles per gallon.
        \end{itemize}

        2. \textbf{Calculate the total miles driven in a year:}
        
        \begin{itemize}[left=10pt,itemsep=0.1em]
        \item The driver travels 4,250 miles per month.
        \item In a year, the driver travels $(4,250 \times 12 = 51,000)$ miles.
        \end{itemize}

        3. \textbf{Calculate the annual fuel consumption:}
        
        \begin{itemize}[left=10pt,itemsep=0.1em]
        \item The fuel consumption is the total miles driven divided by the miles per gallon.
        \item Annual fuel consumption = $\left(\frac{51,000 \text{ miles}}{33 \text{ miles per gallon}}\right)$.
        \end{itemize}

        4. \textbf{Perform the division:}
        
        \begin{itemize}[left=10pt,itemsep=0.1em]
        \item $\left(\frac{51,000}{33} \approx 1,545.45\right)$ gallons.
        \end{itemize}

        5. \textbf{Select the closest option:}
        
        \begin{itemize}[left=10pt,itemsep=0.1em]
        \item The closest option to 1,545.45 gallons is (B) 1,500.
        \end{itemize}

        Therefore, the predicted annual consumption of fuel is $\boxed{B}$. \\ 
        \colorbox{gray!10}{\texttt{</think>}}\colorbox{gray!10}{\texttt{<answer>}} \\
        The answer is $\boxed{B}$ \\
        \colorbox{gray!10}{\texttt{</answer>}}
        \end{tcolorbox}

    \end{vqaexample}
    \label{fig:example_chart}
\end{figure}

\newpage

\begin{figure}[!ht]
    \centering
    \begin{vqaexample}[width=\textwidth]{Logical Reasoning}
        \vspace{2mm}
        \begin{minipage}[t]{0.5\textwidth}
            \centering
            \includegraphics[width=\textwidth]{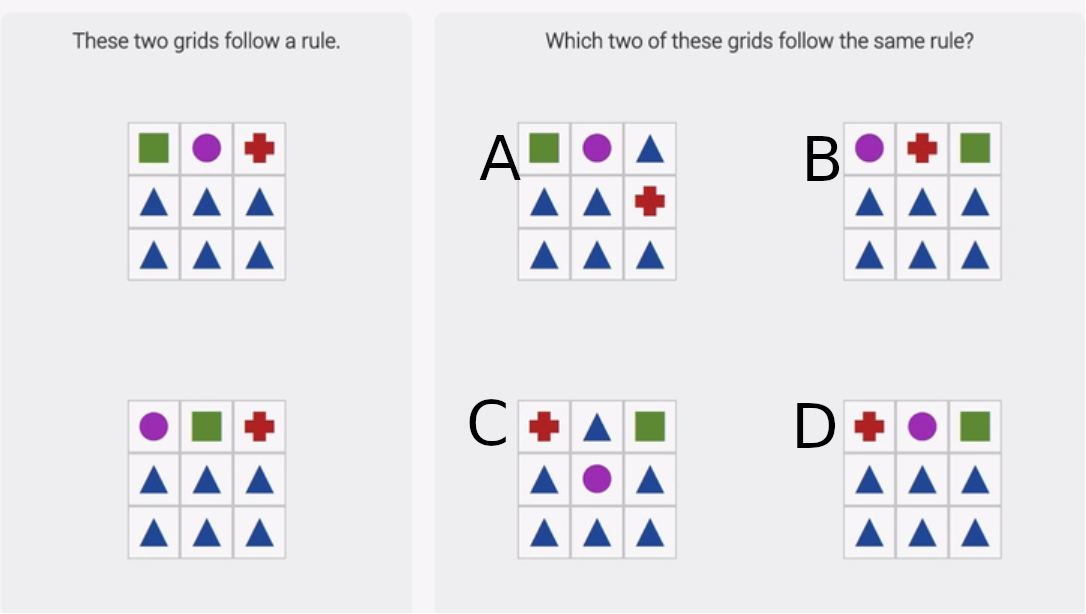}
        \end{minipage}
        \hfill
        \begin{minipage}[t]{0.48\textwidth}
            \begin{tcolorbox}[
                enhanced,
                colback=questionbg,
                colframe=bordercolor!70,
                boxrule=0.8pt,
                arc=3pt,
                left=6pt,
                right=6pt,
                top=4pt,
                bottom=4pt,
                fonttitle=\bfseries\color{titlecolor},
                title=Question: ,
                attach title to upper
            ]
                \ \small{Two grids containing colored symbols and following a common rule are presented. In the block on the right, four additional grids are presented. The candidate must find the two grids that follow the same rule out of these four options. What options (A, B, C, or D) follow this same rule?}
            \end{tcolorbox}
        \end{minipage}
        
        \vspace{0.8em}
        
        \begin{tcolorbox}[
            enhanced,
            breakable,
            colback=answerbg,
            colframe=bordercolor!70,
            boxrule=0.8pt,
            arc=3pt,
            left=6pt,
            right=6pt,
            top=4pt,
            bottom=4pt,
            fonttitle=\bfseries\color{titlecolor},
            title=Answer: ,
            attach title to upper
        ]
            \ \colorbox{gray!10}{\texttt{<think>}}
            
            \textbf{Step 1: Identify the Rule}
            
            The initial two grids are:
            
            1. Grid 1: [Green Square, Purple Circle, Red Cross]
            
            2. Grid 2: [Purple Circle, Green Square, Red Cross]
            
            Both grids have the same sequence of shapes in the bottom row: [Blue Triangle, Blue Triangle, Blue Triangle].
            
            The rule appears to be that the bottom row is always [Blue Triangle, Blue Triangle, Blue Triangle], and the top row contains the same three shapes in a different order. \\
            \\
            \textbf{Step 2: Apply the Rule to the Options}
            
            Now, let's check each option to see if they follow the same rule:
            
            \begin{itemize}[leftmargin=*,itemsep=0.1em]
            \item \textbf{Option A:} [Green Square, Purple Circle, Blue Triangle]
                \begin{itemize}[left=10pt, itemsep=0.1em]
                    \item Bottom row: [Blue Triangle, Blue Triangle, Blue Triangle]
                    \item Top row: [Green Square, Purple Circle, Red Cross]
                    \item  This does not follow the rule because the top row does not contain the same three shapes in a different order.
                \end{itemize}
            \item \textbf{Option B:} [Purple Circle, Red Cross, Green Square]
                \begin{itemize}[left=10pt, itemsep=0.1em]
                    \item Bottom row: [Blue Triangle, Blue Triangle, Blue Triangle]
                    \item Top row: [Purple Circle, Red Cross, Green Square]
                    \item This follows the rule because the bottom row is [Blue Triangle, Blue Triangle, Blue Triangle], and the top row contains the same three shapes in a different order.
                \end{itemize}

            \item \textbf{Option C:} [Red Cross, Blue Triangle, Green Square]
                \begin{itemize}[left=10pt, itemsep=0.1em]
                    \item Bottom row: [Blue Triangle, Purple Circle, Blue Triangle]
                    \item This does not follow the rule because the bottom row is not [Blue Triangle, Blue Triangle, Blue Triangle].
                \end{itemize}

            \item \textbf{Option D:} [Red Cross, Purple Circle, Green Square]
                \begin{itemize}[left=10pt, itemsep=0.1em]
                    \item Bottom row: [Blue Triangle, Blue Triangle, Blue Triangle]
                    \item Top row: [Red Cross, Purple Circle, Green Square]
                    \item  This follows the rule because the bottom row is [Blue Triangle, Blue Triangle, Blue Triangle], and the top row contains the same three shapes in a different order.
                \end{itemize}
            \end{itemize}
            
            \textbf{Conclusion}
            
            The two options that follow the same rule are B and D.
            
            \colorbox{gray!10}{\texttt{</think>}}
        
        \vspace{0.3em}
        \colorbox{gray!10}{\texttt{<answer>}} \\
        The answer is $\boxed{\text{B, D}}$ \\
        \colorbox{gray!10}{\texttt{</answer>}}
        \end{tcolorbox}

    \end{vqaexample}
    \label{fig:example_logic}
\end{figure}

\newpage

\begin{figure}[!ht]
    \centering
    \begin{vqaexample}[width=\textwidth]{Counting Reasoning}
        \begin{minipage}[t]{0.35\textwidth}
            \centering
            \includegraphics[width=\textwidth]{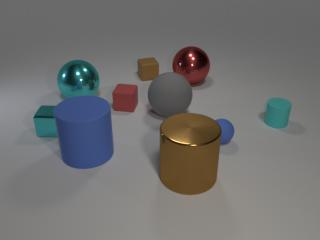}
        \end{minipage}
        \hfill
        \begin{minipage}[t]{0.62\textwidth}
            \begin{tcolorbox}[
                enhanced,
                colback=questionbg,
                colframe=bordercolor!70,
                boxrule=0.8pt,
                arc=3pt,
                left=6pt,
                right=6pt,
                top=4pt,
                bottom=4pt,
                fonttitle=\bfseries\color{titlecolor},
                title=Question: ,
                attach title to upper
            ]
                \ Subtract all tiny yellow matte cylinders. Subtract all small blue rubber spheres. How many objects are left?
            \end{tcolorbox}
        \end{minipage}
        
        \vspace{0.8em}
        
        \begin{tcolorbox}[
            enhanced,
            breakable,
            colback=answerbg,
            colframe=bordercolor!70,
            boxrule=0.8pt,
            arc=3pt,
            left=6pt,
            right=6pt,
            top=4pt,
            bottom=4pt,
            fonttitle=\bfseries\color{titlecolor},
            title=Answer: ,
            attach title to upper
        ]
            \ \colorbox{gray!10}{\texttt{<think>}}
            
            First, let's identify all objects in the image and then apply the given instructions step by step.
            
            \textbf{1. List all objects in the image:}
            \begin{itemize}[left=10pt, itemsep=0.1em]
                \item Large blue matte cylinder
                \item Large gold metallic cylinder
                \item Large gray matte sphere
                \item Small blue rubber sphere
                \item Small cyan rubber cylinder
                \item Small red matte cube
                \item Small brown matte cube
                \item Small cyan metallic cube
                \item Large cyan metallic sphere
                \item Large red metallic sphere
            \end{itemize}
            
            \textbf{2. Subtract all tiny yellow matte cylinders:}
            \begin{itemize}[left=10pt, itemsep=0.1em]
                \item There are no tiny yellow matte cylinders in the image, so no objects are removed.
            \end{itemize}
            
            \textbf{3. Subtract all small blue rubber spheres:}
            \begin{itemize}[left=10pt, itemsep=0.1em]
                \item There is one small blue rubber sphere in the image. Remove this object.
            \end{itemize}
            
            \textbf{4. Count the remaining objects:}
            \begin{itemize}[left=10pt, itemsep=0.1em]
                \item Large blue matte cylinder
                \item Large gold metallic cylinder
                \item Large gray matte sphere
                \item Small cyan rubber cylinder
                \item Small red matte cube
                \item Small brown matte cube
                \item Small cyan metallic cube
                \item Large cyan metallic sphere
                \item Large red metallic sphere
            \end{itemize}
            
            Total remaining objects: 9
            
            The final answer is 9.
            
            \colorbox{gray!10}{\texttt{</think>}}
        
        \vspace{0.3em}
        \colorbox{gray!10}{\texttt{<answer>}} \\
        The answer is $\boxed{9}$ \\
        \colorbox{gray!10}{\texttt{</answer>}}
        \end{tcolorbox}

    \end{vqaexample}
    \label{fig:example_count}
\end{figure}

\clearpage

\begin{figure}[!ht]
    \centering
    \begin{vqaexample}[width=\textwidth]{Diagram Analysis}
        \footnotesize
        \vspace*{-3mm}
        \begin{minipage}[t]{0.15\textwidth}
            \centering
            \includegraphics[width=\textwidth]{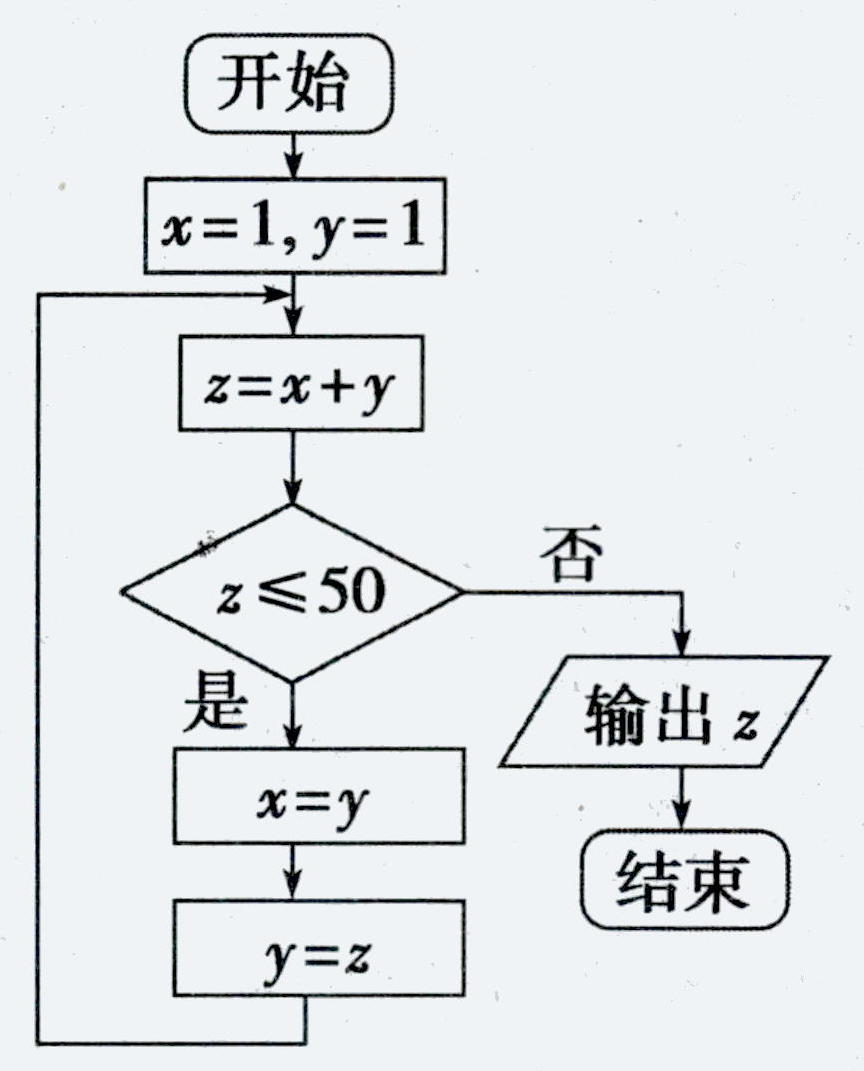}
        \end{minipage}
        \hfill
        \begin{minipage}[t]{0.8\textwidth}
            \begin{tcolorbox}[
                enhanced,
                colback=questionbg,
                colframe=bordercolor!70,
                boxrule=0.8pt,
                arc=3pt,
                left=6pt,
                right=6pt,
                top=4pt,
                bottom=0pt,
                fonttitle=\bfseries\color{titlecolor},
                title=Question: ,
                attach title to upper
            ]
                \ As shown in the figure, the output result of the program flowchart is \_\_\_.
            \end{tcolorbox}
        \end{minipage}
        
        \vspace*{-1mm}
        
        \begin{tcolorbox}[
            enhanced,
            breakable,
            colback=answerbg,
            colframe=bordercolor!70,
            boxrule=0.8pt,
            arc=3pt,
            left=6pt,
            right=6pt,
            top=4pt,
            bottom=4pt,
            fonttitle=\bfseries\color{titlecolor},
            title=Answer: ,
            attach title to upper
        ]
        
            \ \colorbox{gray!10}{\texttt{<think>}}We need to trace the execution of the program flowchart step-by-step until the condition $z > 50$ is met:
            
            \begin{enumerate}[leftmargin=*,itemsep=0.05em]
                \item Start with $x = 1$ and $y = 1$.
                \item Calculate $z = x + y$.
                \item Check if $z \leq 50$. If true, update $x$ and $y$ respectively, and repeat from step 2. If false, output $z$ and end.
            \end{enumerate}
            
            Let's trace through the flowchart:
            
            \begin{itemize}[leftmargin=*,itemsep=0.05em]
                \item \text{Step 1:} $x = 1, y = 1$
                \item Step 2: $z = x + y = 1 + 1 = 2$
                \item Step 3: $z \leq 50$ is true. Update $x = y = 1$ and $y = z = 2$.
            \end{itemize}
            
            Repeat the steps with new $x$ and $y$:
            
            \begin{itemize}[leftmargin=*,itemsep=0.05em]
                \item \text{Step 1:} $x = 1$ and $y = 2$
                \item Step 2: $z = x + y = 1 + 2 = 3$
                \item Step 3: $z \leq 50$ is true. Update $x = y = 2$ and $y = z = 3$.
            \end{itemize}

            Repeat the steps with new $x$ and $y$:
            
            \begin{itemize}[leftmargin=*,itemsep=0.05em]
                \item \text{Step 1:} $x = 2$ and $y = 3$
                \item Step 2: $z = x + y = 2 + 3 = 5$
                \item Step 3: $z \leq 50$ is true. Update $x = y = 3$ and $y = z = 5$.
            \end{itemize}

            Repeat the steps with new $x$ and $y$:
            
            \begin{itemize}[leftmargin=*,itemsep=0.05em]
                \item \text{Step 1:} $x = 3$ and $y = 5$
                \item Step 2: $z = x + y = 3 + 5 = 8$
                \item Step 3: $z \leq 50$ is true. Update $x = y = 5$ and $y = z = 8$.
            \end{itemize}

            Repeat the steps with new $x$ and $y$:
            
            \begin{itemize}[leftmargin=*,itemsep=0.05em]
                \item \text{Step 1:} $x = 5$ and $y = 8$
                \item Step 2: $z = x + y = 5 + 8 = 13$
                \item Step 3: $z \leq 50$ is true. Update $x = y = 8$ and $y = z = 13$.
            \end{itemize}

            Repeat the steps with new $x$ and $y$:
            
            \begin{itemize}[leftmargin=*,itemsep=0.05em]
                \item \text{Step 1:} $x = 8$ and $y = 13$
                \item Step 2: $z = x + y = 8 + 13 = 21$
                \item Step 3: $z \leq 50$ is true. Update $x = y = 13$ and $y = z = 21$.
            \end{itemize}
            
            Repeat the steps with new $x$ and $y$:
            
            \begin{itemize}[leftmargin=*,itemsep=0.05em]
                \item \text{Step 1:} $x = 13$ and $y = 21$
                \item Step 2: $z = x + y = 13 + 21 = 34$
                \item Step 3: $z \leq 50$ is true. Update $x = y = 21$ and $y = z = 34$.
            \end{itemize}
            
            Repeat the steps with new $x$ and $y$:
            
            \begin{itemize}[leftmargin=*,itemsep=0.05em]
                \item \text{Step 1:} $x = 21$ and $y = 34$
                \item Step 2: $z = x + y = 21 + 34 = 55$
                \item Step 3: $z \leq 50$ is false. Output $z = 55$ and end.
            \end{itemize}
            
            The output result is $55$.\colorbox{gray!10}{\texttt{</think>}}
            
            \vspace{0.3em}
            \colorbox{gray!10}{\texttt{<answer>}}The answer is $\boxed{55}$\colorbox{gray!10}{\texttt{</answer>}}
        \end{tcolorbox}
    \end{vqaexample}
    \vspace{-1.5mm}
    \label{fig:example_diagram}
\end{figure}

\end{document}